# When lies are mostly truthful: automated verbal deception detection for embedded lies

**Riccardo Loconte**[1,2,*], **Bennett Kleinberg**[2,3]

[1] Molecular Mind Lab, IMT School of Advanced Studies Lucca, Lucca, Italy

[2] Department of Methodology and Statistics, Tilburg University, Tilburg, Netherlands

[3] Department of Security and Crime Science, University College London, UK

[*] Corresponding author; e-mail: riccardo.loconte@imtlucca.it

**Abstract:**

Background: Verbal deception detection research relies on narratives and commonly assumes statements as truthful or deceptive. A more realistic perspective acknowledges that the veracity of statements exists on a continuum with truthful and deceptive parts being embedded within the same statement. However, research on embedded lies has been lagging behind.
Methods: We collected a novel dataset of 2,088 truthful and deceptive statements with annotated embedded lies. Using a within-subjects design, participants provided a truthful account of an autobiographical event. They then rewrote their statement in a deceptive manner by including embedded lies, which they highlighted afterwards and judged on lie centrality, deceptiveness, and source.
Results: We show that a fined-tuned language model (Llama-3-8B) can classify truthful statements and those containing embedded lies with 64% accuracy. Individual differences, linguistic properties and explainability analysis suggest that the challenge of moving the dial towards embedded lies stems from their resemblance to truthful statements. Typical deceptive statements consisted of 2/3 truthful information and 1/3 embedded lies, largely derived from past personal experiences and with minimal linguistic differences with their truthful counterparts.
Conclusion: We present this dataset as a novel resource to address this challenge and foster research on embedded lies in verbal deception detection.

# Introduction

Everyone engages in some form of deception daily [1]. Rather than fabricating entirely false accounts, however, most individuals tend to combine elements of truth with elements of falsehood [2]. This deception strategy is known as the embedding of lies. Embedded lies present a distinctive challenge in deception research and remain a largely under-investigated phenomenon.

**Verbal deception detection**
Verbal deception detection has a rich history of manual methods developed and refined over decades, which involve training human judges to evaluate statements based on verbal cues. One of the most established techniques is the Criteria-Based Content Analysis (CBCA) [3] originally developed to evaluate children's testimonies on alleged sexual abuse cases and now used to assess the credibility of testimonies in legal contexts. CBCA requires a human to identify and score a narrative on specific verbal cues, such as the amount of detail, unexpected complications, or spontaneous corrections, that truth-tellers are more likely to exhibit than deceivers. Another widely used technique is Reality Monitoring (RM) [4,5], which distinguishes between truth and lies by focusing on the richness of sensory and contextual details provided by the speaker. Truth-tellers are thought to provide more vivid and detailed sensory information than deceivers, who typically rely on fabricated or imagined events. Building on the RM approach, the Verifiability Approach (VA) [6] capitalises on the tendency of truth-tellers to provide more verifiable details compared to lie-tellers, who avoid that because it could expose their deceit. While these methods have promise [3, 7-9], they are more time-consuming and reliant on the expertise of practitioners than automated procedures with computational models, limiting their scalability [10].

**Computer-automated verbal deception detection**
Recent advances in Natural Language Processing (NLP) have introduced automated methods for detecting deception, often combined with methods from machine learning (ML), enhancing both scalability and objectivity.

NLP techniques allow the representation of textual data in a numerical vector form, with different levels of granularity. For instance, the Linguistic Inquiry and Word Count (LIWC) [11] computes the frequency of words that pertain to psychological, social, and emotional dimensions (e.g., cognitive words, affective words, social words, etc); part-of-speech (POS) tagging informs on the shallow syntactic text structure by automatically assigning grammatical categories (e.g., nouns, verbs, pronouns) to words; named-entity recognition (NER) identifies and labels proper nouns and piece of information into predefined broader categories (i.e., *25.12.2024* into DATES, *Central Park* into LOCATIONS, *Google* into ORGANIZATIONS); *n*-gram models represents the text into frequent patterns of tokens; and embeddings use a vectorial numerical representation of words or statements that preserves the semantic and contextual relationship, allowing similar items to have closer representations in a multi-dimensional space.

A common approach for works in computer-automated verbal deception detection (e.g., [12-15]) is to first extract information from textual data and then use supervised machine learning to train models that use these extracted variables to derive a truthfulness judgment (see [10], [16], [17] for an extensive overview on the topic). For example, a previous study detected deceptive opinions by training a naïve Bayes and a support vector machine classifier on n-grams reaching 70.8% and 70.1% accuracy [12]. Likewise, another study detected deceptive opinions after training a support vector machine classifier on a combination of n-grams and LIWC features, reaching 89.8% accuracy [13]. Finally, other scholars extracted the proportion of unique named-entities and found a significant discriminative power of 0.67 and 0.65 in detecting positive and negative deceptive hotels reviews, respectively [14]. Some other works have used recent advances with Transformer-based models [18]. For example, a previous study employed a Bidirectional Encoder Representations for Transformers (BERT) [19] to incorporate contextual embeddings with attention-based mechanisms and detected deceptive utterances in a dataset of transcripts of criminal proceedings hearings, reaching 71.61% accuracy (DeCour) [20]. Another study fine-tuned a large language model (i.e., FLAN-T5) [22] to classify deceptive statements in three datasets encompassing personal opinions, autobiographical events, and future intentions, reaching 79.31% accuracy [22].

This increasing sophistication of NLP techniques paves the way for future research to further refine methods and address more complex challenges in automated deception detection, such as the detection of embedded lies.

**Embedded deception**
The concept of embedded lies emerged when researchers started to ask lie-tellers about their lying strategies. Across a range of different contexts and studies, lie-tellers were found to frequently draw on past experiences to make their lies more believable [23-27]. For instance, in two studies, 67% and 86% of liars, respectively, chose to construct their deceptive statements by incorporating elements of previously experienced events [28]. Similarly, in another research, over 20% of deceptive statements consisted of truthful information [6].

Despite being widely acknowledged within the field of deception detection, only a limited number of studies have directed their attention towards the phenomenon of embedded lies. Most research has focused on fabrication, conceptualising fabricated stories as entirely false and resulting in a dichotomous view of deception as either completely deceptive or not deceptive at all. A study typical of this perspective research requires a between-subjects design where participants engage in or view a mock crime event and are then allocated to one of two conditions: truth-tellers, in which they recollect exactly what they watched or experienced, and lie-tellers, in which they fabricate the event. For example, in a previous study, truth-tellers played a game during a staged event with a confederate, while lie-tellers -who did not participate in the event- were instructed to steal £10 from a wallet and then fabricated their involvement in the staged activities during a subsequent interview [29]. In other experiments, researchers implemented a matched-paired design to match participants with the specific content of false statements [30-32]. For example, participants in the honest condition were asked to tell the researcher about their past holidays, whereas, in the deceptive condition, participants were instructed to pretend to have experienced that same holiday [32].

However, a more nuanced and realistic perspective acknowledges that a deceptive statement exists on a continuum where truthful and deceptive parts are embedded within the same statement (see Fig. 1). We, hence, adopted this framework in this paper to address the challenge of moving the dial towards embedded lies.

**FIGURE 1**
*Graphical representation of the deception continuum framework.*

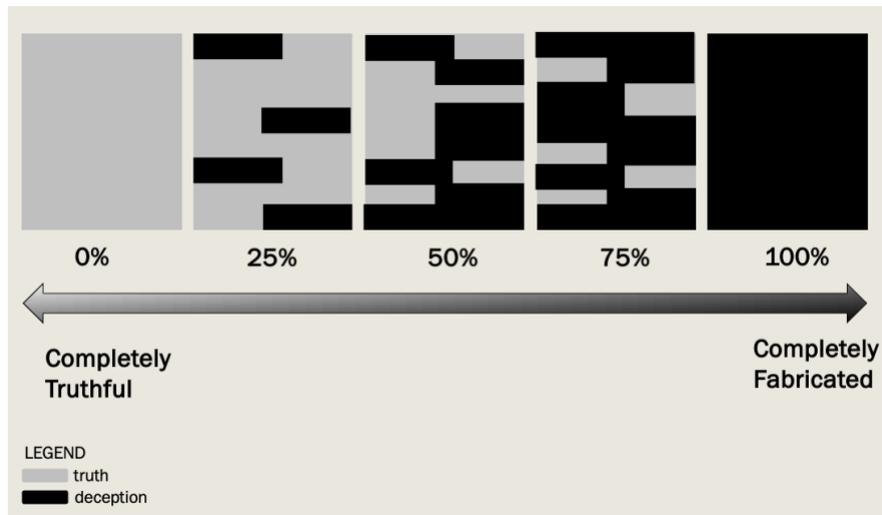

*Note.* Deception is embedded into truthful statements in a continuous from 0% (= no lies are present) to 100% (= the whole statement is made up). Levels in between represent various degrees of deception in the forms of embedded lies.

## Detecting embedded lies

While embedded deception has been acknowledged as a problem by many [6, 23-28], only a few studies reported on the nature and detection of embedded lies.

Two studies investigated embedded lies as fabricated statements contained within otherwise truthful statements [33, 34]. The first found that lies embedded in a fabricated statement were not qualitatively different from lies embedded in an otherwise truthful statement, suggesting that verbal credibility assessment tools may be robust against the embeddings of lies. The second was a follow-up study employing the within-statement baseline comparison [35], which is a strategy based on the idea that deception can be better detected if compared with a truthful baseline. However, the results showed that this strategy was not effective in improving the deception detection rate.

Another work investigated embedded lies to test whether specific verbal cues of deception, such as complications, common knowledge details and self-handicapping strategies, varied across different amounts of lying in statements [36]. The study found only a significant difference in the number of complications, with a larger difference between truthful and outright deceptive statements compared to truthful and embedded lies.

Another work challenged the dichotomy between bald-faced lies and bald-faced truths [2]. It was hypothesized that both truthful and deceptive narratives draw from a common pool of information so that lies and truths are not mutually exclusive but rather coexist within the same accounts. In this study, participants were randomly allocated to the truthful or deceptive condition and were tasked with writing opinions about their friends. They were then instructed to indicate within their statements the specific parts that were deceptive and those that were truthful. The results suggested that truthful narratives inherently include a certain proportion of embedded lies (29.21% of embedded deceptive statements in truthful texts compared to 37.06% in deceptive texts). While deceptive narratives contained a significantly higher rate of fabrications than truthful narratives, it is noteworthy that even the latter was aplenty with fabrications.

The few studies to date have not yet addressed two important challenges in lie detection research. Firstly, measuring embedded lies is inherently complex, with few studies employing within-subjects designs that control for individual differences and statement topics. Secondly, the lack of granular analytical methods hampered the detection of embedded lies, which are harder to identify than general deception. These methodological limitations have hindered the exploration of embedded lies, leaving them underrepresented in the literature despite their significance. Additionally, previous scholars have mentioned the importance of individual differences (e.g., demographic factors, personality traits, cognitive styles, and emotional states) in engaging in deceptive behaviour and in the type and dynamic of the deception involved [37- 44] (for a recent and complete review, see [45]). In the context of embedded lies, only one study explored whether and how personality (i.e., dark triad traits) [46] and demographic factors (i.e., age, gender, ethnicity and political ideology) influence this specific form of deception [2]. However, from this specific study, no significant differences emerged. Hence, with respect to individual differences, embedded lies represent an even more unexplored phenomenon. We, therefore, aim to connect these two streams of research in this paper by also promoting the investigation of individual differences in embedded lies.

**The current study**
This paper aims to help bridge the gap between deception practice and research by focusing explicitly on embedded lies. Prior work has usually employed between-subject or matched-paired designs to study deception intended as fully fabricated accounts. Further, the majority of deception work relies on relatively small datasets [47] and manual procedures (e.g., [48]). Embedded lies need also further investigation in terms of individual differences, with only one study focusing on demographic and individuals' traits affecting embedded deception [2]. We seek to address these limitations. First, we present a new dataset of embedded lies collected in a within-subjects experimental design that is sufficiently large to conduct meaningful computational analysis, including predictive modelling. Second, we enrich the scope of the dataset beyond the narratives and provide data at the individual level, allowing also for analyses of individual differences in verbal deception behaviour. Third, we utilize automated approaches to retrieve variables from the narrative data using NLP methods and further resort to supervised machine learning to train models in detecting embedded deception.

## Materials and Method

**Transparency statement**
This study was approved by the local ERB (Reference Number: anonymised). All data, materials and code to reproduce the analysis are publicly available at https://osf.io/jzrvh/?view_only=0195bd62f6974482b02fbc3c2912dbf4

**Participants**

We recruited a total of 1058 participants[1] fluent in English from the general population through the online participant pool Prolific. Each participant provided informed consent before taking part in the Qualtrics-administered experimental task. Participation in the study was reimbursed 2$ upon experiment completion. Eight participants who did not follow the instructions (i.e., repeated the same phrase in multiple boxes) or provided non-sensical completions to the open-answer fields (e.g., writing random characters to fill the box) were removed for analysis. Eight participants from the subset of participants that freely recall a memorable event (after selecting the option "None of them") were removed because provided a title story that was too long (i.e., with a number of words higher than two standard deviations from the average) and were basically anticipating the main story in the wrong section. The final sample consisted of 1042 participants (58.23% females, 41.17% males, 0.19% preferred not to say, 0.38% expired data or removed consent on Prolific). The mean age was 30.32 years (SD = 9.35, range: 18-105).

**Experimental task**

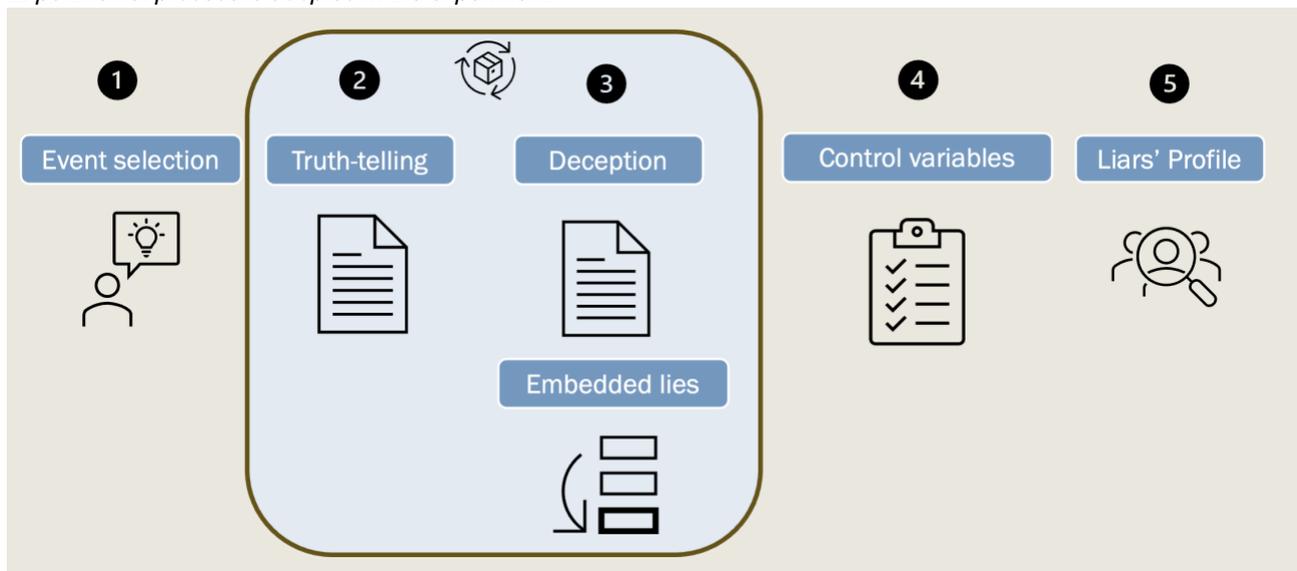

**FIGURE 2**
*Experimental procedure adopted in the experiment.*

*Note.* The order of tasks two and three was counterbalanced across participants.

A previous study found that truthful statements may contain deceptive parts [2]. However, we argue that truthful statements may also be, by definition, completely truthful, and those that are partially deceptive might be residuals from research design limitations.
For this study, we developed an experimental task that followed a different perspective, considering fabrication on a continuum from 0 (fully truthful statements) to 100 (fully deceptive statements).
The experiment was conducted in a within-subjects design (Fig. 2).

---

[1] An *a priori* power analysis for a small effect with a power of 0.90 (Cohen's *d* = 0.20, α = .05, two-tailed) resulted in a sample size of 265 participants. Since we aimed to present a dataset adequate for computational analyses, we collected significantly more data.

Additionally, we collected demographic variables and participants' lying attitudes to advance the understanding of individual differences in deceptive behaviours and challenge the untested assumption that "*all liars are the same*".

**Step 1: Event selection**

The experiment started with the event selection task. Participants were provided with a list of eleven pre-selected autobiographical events that they might have experienced in the past 24 months. The events were deemed relevant for lying in the subsequent deceptive writing task. After participants selected all of the events that they had experienced themselves in the past 24 months, they were randomly assigned to one of them and answered five questions about the event to collect the following memory-related variables:

i) **time**: "*how long ago did the event happen?*" through a multiple-choice question with 25 options (from <1 months to 24 months);

ii) **recollection**: "*how often do you think or talk about this event?*" on a 5-point scale (1=never; 5=always);

iii) **importance**: "*how important is this event to you?*" on a 5-point scale (1=not important at all; 5=extremely important);

iv) **accuracy**: "*how well do you remember this event?*" on a 5-point scale (1=not well at all; 5=extremely well);

v) **valence**: "*how would rate this event in emotional terms?*" on a 5-point scale (-1=extremely negative; 1=extremely positive).

The assigned event served as the basis for the remainder of the task. If participants did not experience any of the events in the list, they were instructed to choose the option "*none of them*". They were then asked to think about a positive or negative event occurring in the last 24 months that was memorable, emotional and that directly involved them and were asked to provide a short title for the event. The 19.52% (n=205) of participants chose the "*none of them*" option.

We focused on autobiographical events, that were deemed relevant for lying, to mirror real-world scenarios and enhance the practical application of our research findings in improving the accuracy of deception detection.

**Step 2: Truth-telling task**

The order of the truth-telling and deception tasks was randomized across participants. In the truth-telling task, participants were required to write an accurate and truthful account of the event in question. They were also asked to use correct spelling and grammar and were reminded not to use AI assistants in the writing task. Copy-pasting was blocked to prevent it. This task required a minimum number of 3000 characters (ca. 428-750 words) to move to the next phase of the experiment. Full instructions are provided in Supplementary Materials (SM) 1.

**Step 3a: Deception task**

For the deception task, participants were provided with a context relevant to lying and instructed to write a deceptive account of the selected event by incorporating false information. Specifically, participants were told to write an alternative version of the selected event in order to get a specific advantage from lying. Participants were also warned to not make up a statement about a new event and to not mention in any way they were lying.

The list of the contexts matched for each event, and the number of participants allocated is provided in Table 1. Only 18.91% of participants (n = 197) did not experience any of the predefined events and were free to recollect a memorable event that occurred in the past 24 months. Other than the deceptive instructions, the general writing instructions were identical to the truthful task. To motivate participants to do their best, they were informed about the chance of winning an extra 50£ if their statement was considered credible by the experts. In reality, all participants were included in the draw and the payment was distributed to a randomly selected participant after the data collection concluded. Full instructions are provided in SM 1.

**TABLE 1**
*List of events, contexts for lying, and number (percentages) of participants allocated to that event.*

| Events | Context for lying | No. of participants allocated (%) |
|---|---|---|
| A job interview for your dream job | Inflate your past experiences to get the job | 160 (15.36%) |
| Being hospitalized and undergoing surgery | Exaggerate some side effects to receive extra compensation from your health insurance | 70 (6.67%) |
| Being involved in a car accident | Increase the claimed amount of damage you received to get more money | 47 (4.51%) |
| Causing a car accident | Describe the event so that it's not your fault | 15 (1.44%) |
| Cheating on an exam | Describing how you passed the exam, given that you cannot admit that you cheated | 48 (4.61%) |
| Cheating on your partner | Convince your partner that you didn't cheat on them | 36 (3.45%) |
| Ending a long romantic relationship | Pretend that you just had an argument with your partner | 152 (14.59%) |
| Getting a speeding fine | Pretend it wasn't you driving the car that day | 62 (5.95%) |
| Getting fired | Pretend that you just had a bad day at work | 34 (3.26%) |
| Missing a deadline at work because of bad organization | Find excuses that allow you not to appear forgetful or disorganised | 97 (9.31%) |
| None of them | - | 197 (18.91%) |
| Taking the bus/train without the ticket | convince the ticket inspector that they shouldn't fine you for not having the ticket | 124 (11.90%) |

**Step 3b: Embedded lie selection**

Once participants had written the deceptive account, participants were instructed to copy and paste words, phrases or sentences from their deceptive statements into a maximum of 20 text boxes (similar to [2]). For each word or phrase that was copy-pasted, participants rated the deceptiveness and centrality of each embedded lie on a 5-point scale (1=not at all deceptive/central to 5=extremely deceptive/central).

Through a multiple-choice question, participants provided the source on which they relied for the embedded lie. The source options were based on liars' relying on their past experiences or cognitive processes (i.e., from memory, imagination, and planning). The following source options were provided: 1) you connected the detail to a past personal experience; 2) you saw a similar event happen to someone else and used that as a basis for the detail; 3) you derived the detail from a story another person told you, or from a book, or a movie; 4) you imagined the detail without any specific memory or experience; 5) you used planned, future activities as a reference.

To account for individual variability (i.e., participants copy-pasting a single word vs. multiple phrases or sentences), for each subject, the number of embedded lies was also standardized by computing the ratio between the number of words provided in the 20 boxes and the total number of words in their deceptive text. The standardized number of embedded lies ranged from 0 to 1.

**Step 4: Control variables**

Once the two writing tasks were completed, participants rated the following control variables on a 5-point scale (1=completely disagree; 5=completely agree): i) difficulty of the task (i.e., "*I found the* *task was difficult*"); ii) clarity of instructions (i.e., "*I found the instructions were clear*"); iii) motivation of telling the truth (i.e., "*I was motivated to provide a convincing truthful statement*"); iv) motivation of lying (i.e., "*I was motivated to provide a convincing deceptive statement*").

**Step 5: Liars' profile**

To measure potential individual differences in participants' lying attitudes, the lying profile questionnaire [49] was administered. The lying profile questionnaire measured dispositional traits of deception and was composed of 16 items grouped into four factors: frequency of lying (frequency); ability to lie (ability); negative attitude towards lying (negativity); and positive attitudes toward lying depending on the context (contextuality). Since participants may be prone to mask their lying attitude, the Balanced Inventory of Socially Desirable Responding Short Form (BIDR) [50] was also administered and used to correct the lying profile scores for potential effects of social desirability. The BIDR was a 16-item questionnaire which measured two main dimensions of social desirability: 1) self-deception enhancement (SDE): the unconscious tendency of individuals to provide honest but positively biased self-reports to protect self-esteem; 2) impression management (IM): the habitual and conscious tendency of individuals to present themselves of a favourable public image. We report results on both the raw lying profile scores as well as the ones after correcting for the BIDR scores. The correction procedure was conducted using four general linear models, fitted to regress out the SDE and IM scores from each lying profile factor.

**Textual analysis of narrative data**

**Linguistic Inquiry and Word Count analysis**

The Linguistic Inquiry and Word Count (LIWC) [11,51] software is the gold standard for analysing word usage and semantics in texts across more than 100 features by calculating the percentage of total words corresponding to each category using validated dictionaries of words associated with psychosocial dimensions. Specifically, the English dictionary (version LIWC-22) was employed for this analysis, and 118 features were extracted from tokenized text.

**DeCLaRatiVE stylometry**

The DeCLaRatiVE stylometry approach [22] subsumes 26 linguistic variables derived from four theoretical lines in verbal deception research: Distancing [52], Cognitive Load [53, 54], Reality Monitoring [4, 55], and VErifiability Approach [7, 56]. Linguistic variables associated with the CL, such as text length, readability, and complexity, were computed using the Python library TEXTSTAT. Those related to the Distancing and RM framework were computed using LIWC-22 features [11,51] extracted from tokenized text. RM was also investigated through linguistic concreteness by cross-referencing an annotated dataset [57] with the content words in our dataset and averaging the final scores per statement. The preprocessing steps to derive content words from statements were tokenization, conversion to lowercase, stop-word removal, and lemmatization and were run with the SpaCy library in Python. Finally, verifiable details were extracted as entities with the named-entity recognition (NER) model available on the SpaCy library (en_core_web_trf, https://spacy.io/models/en#en_core_web_trf ). A full list of the 26 linguistic variables with a short description is shown in Table 2 (refer to the original work for a deeper understanding of the approach).

**TABLE 2**
*List and short description of the 26 linguistic features pertaining to the DeCLaRatiVE Stylometry technique.*

| Label | Description |
| --- | --- |
| num_sentences | Total number of sentences |
| num_words | Total number of words |
| num_syllables | Total number of syllables |
| avg_syllabes_per_word | Average number of syllables per word |
| fk_grade | Index of the grade level required to understand the text |
| fk_read | Index of the readability of the text |
| Analytic | LIWC summary statistic analyzing the style of the text in term of analytical thinking (0 - 100) |
| Authentic | LIWC summary statistic analyzing the style of the text in term of authenticity (0 - 100) |
| Tone | Standardized difference (0-100) of 'tone_pos' - 'tone_neg' |
| tone_pos | Percentage of words related to a positive sentiment (LIWC dictionary) |
| tone_neg | Percentage of words related to a negative sentiment (LIWC dictionary) |
| Cognition | Percentage of words related to semantic domains of cognitive processes (LIWC dictionary) |
| memory | Percentage of words related to semantic domains of memory/forgetting (LIWC dictionary) |
| focuspast | Percentage of verbs and adverbs related to the past (LIWC dictionary) |
| focuspresent | Percentage of verbs and adverbs related to the present (LIWC dictionary) |
| focusfuture | Percentage of verbs and adverbs related to the future (LIWC dictionary) |
| Self-reference | Sum of LIWC categories 'i' + 'we' |
| Other-reference | Sum of LIWC categories 'shehe' + 'they' + 'you' |
| Perceptual details | Sum of LIWC categories 'attention' + 'visual' + 'auditory' + 'feeling' |
| Contextual Embedding | Sum of LIWC categories 'space' + 'motion' + 'time' |
| Reality Monitoring | Sum of Perceptual details + Contextual Embedding + Affect - Cognition |
| Concreteness score | Mean of concreteness score of words |
| People | Unique named-entities related to people: e.g., 'Mary', 'Paul', 'Adam' |
| Temporal details | Unique named-entities related to time: e.g., 'Monday', '2:30 PM', 'Christmas' |
| Spatial details | Unique named-entities related to space: e.g., 'airport', 'Tokyo', 'Central park' |
| Quantity details | Unique named-entities related to quantities: e.g., '20%', '5 $', 'first', 'ten', '100 meters' |

**n-gram differentiation**

Using the *n*-gram differentiation test [58] we compared the frequencies of unigrams, bigrams, and trigrams in truthful and deceptive statements within each event. This comparison was made using a signed rank sum test approach. Ties in ranks were fixed by averaging random ranks in 500 iterations. Statements were first pre-processed using spaCy library in Python by removing stop words and lemmatising the remaining words. Only *n*-grams that appeared in at least 5% of all documents were included in the analysis. The effect size used for the frequency comparisons was *r*, which ranged from -1.0 to 1.0.

**Machine-learning classification**

To investigate whether deceptive statements with embedded lies can be distinguished from truthful statements, we performed a document classification task using different state-of-the-art ML approaches. The models included both traditional and advanced architectures. Specifically, four Random Forest (RF) classifiers were trained on Bag of Words (BOW) [59]; representations, LIWC variables [11], DeCLaRatiVE variables [22], and GPT-embedding representations[2], respectively. Additionally, we tested the performance of different fine-tuned language models, such as distilBERT [60], FLAN-T5 base [21], and Llama-3-8B [61]. We finally explored the performance of a deception language model from a previous study [22], which was a FLAN-T5 base model fine-tuned on three large datasets of deception with 79.31% (± 1.3) accuracy.

Language models (i.e., distilBERT, FLAN-T5, and Llama-3) were trained using the HuggingFace library and the Google Colaboratory Pro + interface with the A100 Tensor Core GPU. Cross-validation was performed to ensure robust evaluation. Specifically, RF models were trained using 10-fold nested cross-validation, while language models were fine-tuned with 5-fold cross-validation to optimize computational costs. Classification performance was assessed in terms of overall accuracy, as well as precision, recall, and F1-score by condition (truthful vs deceptive). Details of each model and the training procedure are reported in SM 4.

**Analysis Plan**

We first looked at the subject level to examine characteristics of the reported embedded lies, such as their frequency, source, deceptiveness and centrality for a deceptive account. Second, state-of-the-art machine learning approaches were employed in a classification task to differentiate truthful from deceptive statements with embedded lies. Furthermore, we assessed linguistic differences between the narratives by using the LIWC and DeCLaRatiVE approach [22]. From the LIWC, we obtained for each subject 118 variables; from the DeCLaRatiVE stylometry technique, we obtained 26 variables. A within-subject permutation t-test with 9,999 permutations [62] was employed to test for statistical differences in these variables by statement veracity (truthful vs deceptive). Results from multiple comparisons were corrected using Bonferroni correction. Truthful and deceptive statements were also analysed in terms of n-grams by using the n-grams differentiation test. Finally, we examined individual differences related to demographic variables and lying profiles. Analyses were conducted in R using the *MKinfer* and *effectsize* libraries.

# Results

**Corpus descriptives**

We collected a corpus of 2084 truthful and deceptive statements, collectively, across 11 events deemed relevant for lying[3]. We found deceptive statements (M = 7.13, SD = 4.65) containing a significantly higher average number of sentences than truthful statements (*M* = 6.77, *SD* = 4.48), $t_{(9999)}$ = -0.37, $p < .01$, $d$ = -0.08 [-0.14, -0.03]. Likewise, the number of words was, on average, significantly higher in deceptive (*M* = 145.29, *SD* = 83.65) than in truthful statements (*M* = 131.74, *SD* = 78.49), $t_{(9999)}$ = -13.55, $p < .01$, $d$ = -0.17 [-0.22, -0.12].

---

[2] https://platform.openai.com/docs/guides/embeddings
[3] Descriptive statistics of the variables associated with the events are reported in Table 3S in SM 2.

**Embedded lies**

Embedded lies included an average of 5.03 lies per text (*SD* = 3, *Median* = 4), with an average number of words of 46.27 (*SD* = 42, *Median* = 35, see Table 3). The average ratio between the number of words in the annotated embedded lies and the respective deceptive statement was 0.32 (*SD* = 0.20, Median = 0.29). Embedded lies were perceived as moderately deceptive (*M* = 3.94, *SD* = 0.79, *Median* = 4) and central to the overall narrative (*M* = 3.55, *SD* = 0.82, *Median* = 3.59). Further, 35.86% of embedded lies (*n* = 1881) relied on personal past experiences that involved participants directly and 10.41% (*n* = 546) indirectly; 33.86% of embedded lies (*n* = 1776) relied on participants' imagination, while 14.95% (*n* = 784) on others' experiences and only 4.92% (*n* = 258) on personal future plans. An example of subjects' responses is provided in Box 1. Correlational analysis between variables associated with embedded lies, lying profile scales and BIDR scales are provided in SM 2.

**TABLE 3**
*Descriptive statistics of participants' responses in variables associated with embedded lies (M, SD, Median).*

|  | Embedded lies | | |
| --- | --- | --- | --- |
|  | *M* | *SD* | *Median* |
| Words | 46.27 | 42.23 | 35 |
| Absolute no. of embedded lies | 5.03 | 3.35 | 4 |
| Standardized no. of embedded lies | 0.32 | 0.20 | 0.29 |
| Deceptiveness | 3.94 | 0.79 | 4 |
| Centrality | 3.55 | 0.82 | 3.59 |

**BOX 1**
*Example of a statement provided by participants during the task.*

| EVENT: Being involved in a car accident | EVENT: Being involved in a car accident<br>INSTRUCTIONS: lie about the event to increase the claimed amount of damage you received to get more money |
| --- | --- |
| «I was driving home after getting my dog from her sitter. My mom was sitting next to me to keep company to the dog, when we got met with a lot of traffic. So we were advancing quite slowly towards our destination when we come across this intersection, where on the right the cars have a STOP sign. This guy, very old, probably in his 60s, doesn't stop and continues moving towards us. I stomped on the break, but it wasn't in time, and the car crashed against our side. It had been years since I was involved in anything of the sort, so while I was pretty sure it was not my fault, I was shaking the entire time I was dealing with the men to fill out the paperwork». | «I was driving home after I got my dog from her sitter. My mom and dog were sitting in the passenger seat, **my dog likes to ride on the ground between my moms legs**. We came across quite a bit of traffic and were moving slowly towards our destination. As I move through this intersection, where cars on the right have a STOP sign (so they have to stop, and I have priority), this guy thats at least 60 years old, completely ignores the sign and advances towards us at quite a speed. Because there was traffic in front of me, **I could nothing but watch** as the car crashed into us, directly on the passengers side. **My mom was thrown to my side**, kept in place only by the seat belt, and **her leg was pretty badly hurt**, cause **she used her body to protect our doggy**. This experience is clear in my mind because after I got off **I had to have a fight with the other driver because he was incapable of acknowledging fault**». |

*Note.* On the left side, a sample truthful statement from a participant telling when being involved in a car accident. On the right side, the same participant providing the deceptive statements about the same event following the given instructions. In bold the embedded lies identified by the participant.

**Predictive modelling performance**
We trained different machine learning and language models to distinguish deceptive statements with embedded lies from truthful ones. Table 4 shows that all models could classify statements better than the chance level (with p < .01 after running an exact binomial test), but the highest performance reached 64% accuracy after fine-tuning a Llama-3 model.

**TABLE 4**
*Classification performance of predictive models.*

| Model | Accuracy | Truthful | | | Deceptive | | |
|---|---|---|---|---|---|---|---|
| | | Precision | Recall | F1 | Precision | Recall | F1 |
| BOW + RF | .55 (.03) | .55 (.03) | .53 (.04) | .54 (.03) | .55 (.03) | .56 (.04) | .55 (.03) |
| LIWC + RF | .58 (.02) | .58 (.02) | .58 (.05) | .58 (.03) | .58 (.02) | .57 (.05) | .57 (.03) |
| DeCLaRatiVE + RF | .58 (.03) | .59 (.04) | .56 (.05) | .57 (.04) | .58 (.03) | .60 (.06) | .59 (.04) |
| GPT-embeddings + RF | .62 (.03) | .62 (.03) | .62 (.05) | .62 (.03) | .62 (.03) | .62 (.05) | .62 (.03) |
| distilBERT | .60 (.02) | .64 (.05) | .51 (.19) | .55 (.10) | .60 (.05) | .69 (.16) | .63 (.05) |
| Fine-tuned FLAN-T5 base | .60 (.02) | .60 (.06) | .57 (.03) | .59 (.03) | .59 (.04) | .63 (.04) | .61 (.01) |
| **Fine-tuned Llama-3-8B** | **.64 (.04)** | **.67 (.05)** | **.55 (.13)** | **.60 (.08)** | **.62 (.05)** | **.73 (.10)** | **.67 (.05)** |
| Deception language model | .56 | .54 | .76 | .63 | .60 | .35 | .44 |

*Note.* The values refer to the average performance after performing cross-validation. In brackets, the standard deviation is reported. The deception language model was only employed to predict the class in our dataset; therefore, no cross-validation was performed. All models were significantly better than the chance level with p < .01. In bold is the performance of the best model.
[a] BOW = bag of words
[b] RF = random forest
[c] LIWC = Linguistic inquiry and word count

**Exploratory explainability analysis**

To add interpretations on the achieved performance, we conducted an explainability analysis on the Llama-3 and deception language model. We computed Spearman's rank correlations between the deceptive class probabilities and the absolute and standardized number of embedded lies, deceptiveness and centrality scores (Table 5S in SM 3). There was a significant positive correlation between the class probability of deceptiveness and the absolute (*rho* = .10, *S* = 170216978, *p* < .01) and standardized number of embedded lies (*rho* = .10, *S* = 170565831, *p* = .001). For the deception language model, we found a significant positive correlation between the absolute number of embedded lies and class probability (*rho* = .09, *S* = 171758230, *p* = .004). Finally, only for the Llama-3 model, we found correct classifications having a significantly higher amount of absolute number of embedded lies (M = 5.31, SD = 3.39) compared to incorrect ones (M = 4.43, SD = 2.83), d = 0.27 [0.14, 0.40]. Similarly, a standardized number of embedded lies was significantly higher in correctly classified statements (M = 0.34, SD = 0.21) with respect to incorrect ones (M = 0.29, SD = 0.19), d = 0.22 [0.09, 0.35] (see Table 6S in SM 3). These findings suggest that the more a statement is fabricated, namely, the greater the number of embedded lies within an otherwise truthful statement, the higher the probability of a language model to accurately and confidently predict the class of that statement.

**Textual analysis of narrative data**

Tables 5 and 6 suggest that a few linguistic indicators were significantly indicative of deception, albeit often with small effect sizes. LIWC variables associated with truthful statements pertained mainly to using social words and references (i.e., social words, social references, pronouns and personal pronouns, social behaviour, language of status and leadership; Table 5). In contrast, LIWC features associated with deception included mainly words associated with cognitive processes (i.e., words related to memory, cognition, and differentiation). Usage of emotional words (i.e., anger, sadness) and text statistics (e.g., word counts, number of periods) showed ambiguous patterns, with some features more associated with truthful statements and others more associated with deceptive statements.

When we conducted the analysis by event, significant differences emerged for some LIWC variables by statement veracity for four events (i.e., Being hospitalized and undergoing surgery, Ending a long romantic relationship, Getting a speeding fine, and Taking the bus/train without the ticket).

**TABLE 5**
*Effect sizes (and CIs) of significant LIWC features for the entire dataset and specific events.*

| Topic | LIWC feature | LIWC Interpretation | LIWC example words | Cohen's d | Adjusted CI | Direction |
|---|---|---|---|---|---|---|
| Overall | Social | Social words | Argue, boyfriend, chat | 0.26 | 0.13, 0.38 | T > D |
| | socrefs | Social references | you, we, he, she | 0.23 | 0.11, 0.35 | T > D |
| | shehe | Third singular personal pronouns | she, he, her, his | 0.23 | 0.10, 0.35 | T > D |
| | WC | Total word counts | - | 0.22 | 0.10, 0.34 | T > D |
| | ppron | Personal pronouns | I, you, my, me | 0.21 | 0.09, 0.33 | T > D |
| | Period | Number of sentences | - | -0.19 | -0.31, -0.07 | D > T |
| | male | Male references | he, his, him, man | 0.18 | 0.06, 0.31 | T > D |
| | memory | Memory words | remember, forget, remind | -0.17 | -0.29, -0.05 | D > T |
| | Clout | Language of leadership, status | - | 0.16 | 0.04, 0.28 | T > D |
| | socbehav | Social behavior words | said, love, say, care | 0.15 | 0.03, 0.27 | T > D |
| | differ | Words of differentiation | but, not, if, or | -0.14 | -0.27, -0.02 | D > T |
| | emo_anger | Emotion of anger | hate, mad, angry, frustr* | 0.14 | 0.02, 0.26 | T > D |
| | pronoun | Pronouns | I, you, that, it | 0.14 | 0.02, 0.26 | T > D |
| | emo_sad | Emotion of sadness | sad, disappoint*, cry | -0.14 | -0.26, -0.01 | D > T |
| | number | Numbers | one, two, first, once | -0.13 | -0.25, -0.01 | D > T |
| | Cognition | Words of cognition | know, think, but, if | -0.12 | -0.25, -0.003 | D > T |
| Being hospitalized and undergoing surgery | Tone | Emotional tone | - | -0.49 | -0.94, -0.04 | D > T |
| | Period | Number of sentences | - | -0.48 | -0.92, -0.03 | D > T |
| | WC | Total word count | - | 0.47 | 0.02, 0.92 | T > D |
| | power | Words of power | own, order, allow, power | 0.45 | 0.002, 0.89 | T > D |
| Ending a long romantic relationship | emo_anger | Emotion of anger | hate, mad, angry, frustr* | 0.41 | 0.12, 0.71 | T > D |
| | conflict | Conflict words | fight, kill, killed, attack | 0.36 | 0.06, 0.66 | T > D |
| | ppron | Personal pronouns | I, you, my, me | 0.34 | 0.04, 0.63 | T > D |
| | socbehav | Social behavior words | said, love, say, care | 0.32 | 0.02, 0.61 | T > D |
| Getting a speeding fine | Social | Social words | Argue, boyfriend, chat | 0.75 | 0.24, 1.26 | T > D |
| | socrefs | Social references | you, we, he, she | 0.70 | 0.19, 1.20 | T > D |
| | shehe | Third singular personal pronouns | she, he, her, his | 0.68 | 0.18, 1.18 | T > D |
| | Clout | Language of leadership, status | - | 0.56 | 0.07, 104 | T > D |
| Taking the bus/train without the ticket | Social | Social words | Argue, boyfriend, chat | 0.65 | 0.30, 1.00 | T > D |
| | shehe | Third singular personal pronouns | she, he, her, his | 0.57 | 0.23, 0.92 | T > D |
| | socrefs | Social references | you, we, he, she | 0.56 | 0.22, 0.90 | T > D |
| | male | Male references | he, his, him, man | 0.54 | 0.20, 0.88 | T > D |

| Topic | LIWC feature | LIWC Interpretation | LIWC example words | Cohen's d | Adjusted CI | Direction |
|---|---|---|---|---|---|---|
| | socbehav | Social behavior words | said, love, say, care | 0.50 | 0.16, 0.84 | T > D |
| | comm | Communication words | said, say, tell, thank* | 0.46 | 0.13, 0.79 | T > D |
| | ppron | Personal pronouns | I, you, my, me | 0.41 | 0.08, 0.74 | T > D |
| | pronoun | Pronouns | I, you, that, it | 0.41 | 0.07, 0.74 | T > D |
| | Cognition | Words of Cognition | know, think, but, if | -0.38 | -0.70, -0.04 | D > T |
| | WC | Total word count | - | 0.35 | 0.03, 0.68 | T > D |
| | tentat | Words of tentativeness | if, or, any, something | -0.34 | -0.67, -0.01 | D > T |

*Note.* Confidence intervals are adjusted for multiple comparisons using Bonferroni correction. Linguistic features are sorted by the absolute value of the effect size magnitude for each event. For the direction of the effect, T = truthful and D = deceptive.

For DeCLaRatiVE linguistic features (Table 6), only a few of them were significantly indicative of deception in nine out of eleven events. When testing the whole dataset, the only significant features were number of words and number of syllables. This finding might be an artefact of the instructions, where participants were instructed to add details in order to appear deceptive and achieve a goal, hence resulting in producing longer statements with more complex words.

**TABLE 6**
*Effect sizes (and CIs) of significant DeCLaRatiVE features for the entire dataset and specific events.*

| Topic | DeCLaRatiVE feature | Cohen's d | Adjusted CI | Direction |
|---|---|---|---|---|
| Overall | num_words | -0.22 | -0.33, -0.11 | D > T |
| | num_syllables | -0.22 | -0.33, -0.11 | D > T |
| A job interview for your dream job | Tone | -0.52 | -0.78, -0.26 | D > T |
| | tone_neg | 0.47 | 0.21, 0.73 | T > D |
| | Contextual Embedding | 0.44 | 0.18, 0.70 | T > D |
| | Other-reference | 0.34 | 0.09, 0.59 | T > D |
| | Analytic | -0.33 | -0.58, -0.07 | D > T |
| | tone_pos | -0.28 | -0.53, -0.03 | D > T |
| Being hospitalized and undergoing surgery | Tone | 0.49 | 0.09, 0.88 | T > D |
| | num_words | -0.47 | -0.86, -0.08 | D > T |
| | num_syllables | -0.47 | -0.86, -0.07 | T > D |
| Being involved in a car accident | Tone | 0.61 | 0.11, 1.11 | T > D |
| | tone_pos | 0.56 | 0.07, 1,05 | T > D |
| | Contextual_Embedding | -0.55 | -1.04, -0.06 | D > T |
| | Reality Monitoring | -0.50 | -0.98, -0.01 | D > T |
| Causing a car accident | Reality Monitoring | -1.48 | -2.69, -0.30 | D > T |
| | Contextual_Embedding | -1.29 | -2.42, -0.18 | D > T |
| Cheating on an exam | Other-reference | 0.53 | 0.05, 1.01 | T > D |
| | Reality Monitoring | 0.52 | 0.04, 1.01 | T > D |
| Ending a long romantic relationship | Other-reference | -0.61 | -0.88, -0.33 | D > T |
| | Analytic | 0.45 | 0.18, 0.71 | T > D |
| | tone_neg | -0.42 | -0.69, -0.16 | D > T |
| | Cognition | -0.38 | -0.64, -0.12 | D > T |
| Getting a speeding fine | Contextual Embedding | -0.80 | -1.25, -0.34 | D > T |
| | Reality Monitoring | -0.62 | -1.05, -0.18 | D > T |
| Missing a deadline at work because of bad organization | tone_pos | 0.40 | 0.07, 0.73 | T > D |
| | Tone | 0.34 | 0.34, 0.02 | T > D |
| | focusfuture | -0.34 | -0.66, -0.1 | D > T |
| Taking the bus without the train ticket | tone_pos | 0.44 | 0.15, 0.73 | T > D |
| | num_syllables | -0.36 | -0.64, -0.07 | D > T |

| Topic | DeCLaRatiVE feature | Cohen's d | Adjusted CI | Direction |
|---|---|---|---|---|
| | num_words | -0.35 | -0.64, -0.06 | D > T |
| | tone_neg | 0.29 | 0.005, 0.58 | T > D |

*Note.* Confidence intervals are adjusted for multiple comparisons using Bonferroni correction. Linguistic features are sorted by the absolute value of the effect size magnitude for each event. For the direction of the effect, T = truthful and D = deceptive.

Finally, the *n*-gram differentiation analysis (Table 7) revealed how deceptive statements with embedded lies may appear very similar to their truthful counterparts, resulting in few or no significant differences in word usage. This result highlights the reasons why detecting embedded lies is a hard task.

**TABLE 7**
*Effect sizes (r) and CIs of significant n-grams for specific events after using the n-grams differentiation test.*

| Event | *n*-gram | r | Adjusted CI | Direction |
|---|---|---|---|---|
| Taking the bus/train without the train ticket | tell | -0.20 | -0.37, -0.02 | D > T |
| | ticket | -0.18 | -0.23, -0.14 | D > T |
| | time | -0.14 | -0.26, -0.01 | D > T |
| Ending a long romantic relationship | relationship | -0.07 | 0.001, 0.13 | T > D |
| Missing a deadline at work because of bad organisation | time | 0.10 | 0.004, 0.20 | T > D |
| Cheating on your partner | feel | 0.33 | 0.06, 0.60 | T > D |
| Being hospitalized and undergoing surgery | pain | -0.22 | -0.38, -0.06 | D > T |
| | surgery | -0.14 | -0.22, -0.05 | D > T |
| Getting fired | fire | 0.25 | 0.09, 0.40 | T > D |
| Getting a speeding fine | speed | 0.10 | 0.005, 0.20 | T > D |
| Cheating on an exam | study | -0.28 | -0.43, -0.13 | D > T |
| | answer | 0.19 | 0.01, 0.36 | T > D |
| Causing a car accident | drive | -0.21 | -0.38, -0.03 | D > T |

*Note.* Confidence intervals are adjusted for multiple comparisons using Bonferroni correction. *N*-grams are sorted by effect size after comparing truthful and deceptive statements for each event. For the direction of the effect, T = truthful and D = deceptive.

**Individual differences**

We investigated individual differences in the dependent variables associated with the absolute and standardized number of embedded lies, deceptiveness, and centrality scores. Regarding demographic factors (see also SM 2), we found a gender difference for the average deceptiveness scores (*diff* = 0.11 ± 0.05, *p* = 0.03, *d* = 0.14 [0.02, 0.26]), with females (*M* = 3.98, *SD* = 0.79) reporting higher values than males (*M* = 3.88, *SD* = 0.77). As for age, we found a small significant positive correlation between age and deceptiveness (*rho* = 0.075, *S* = 172823388, *p* = 0.015).

Furthermore, we investigated the presence of subpopulations of liars by a cluster analysis of participants' scores in the four-factor lying profile questionnaire [49]. Lying profile scores were first adjusted for social desirability[4]. We then followed the procedure in Makowski et al., (2021) [49] to cluster participants (see SM 4). Our dataset was deemed suitable for clustering (Hopkins' $H$ = 0.25)[5]. The method agreement procedure supported the existence of two clusters, as indicated by ten methods out of 29 (34.48%). After applying the k-means clustering algorithm, the two clusters accounted for 31.97% of the total variance of the original data. The first cluster (44.72% of the sample) was characterized by participants with very low reported lying ability, low levels of frequency and contextuality, and strong negative attitudes towards lying; the second cluster (55.28% of the sample) was characterized by people with higher levels of contextuality and frequency of deception, very high levels of ability and low levels of negative attitudes towards lying (Fig. 3). Following the original work [51], we labelled the first cluster as the *virtuous* and the second as the *trickster* cluster. To test the validity of this two-cluster solution we trained a logistic regression that used, as features, the adjusted scores of the four scales of the lying profile questionnaire and, as a predicted variable, the labels obtained from the cluster analyses (as in [63]). We obtained an almost perfect classification (accuracy = 0.99). This result supported the validity of our two-cluster solution, confirming that the labels associated with each participant were not randomly assigned but actually reflected an inherently different pattern of responding. However, no significant differences were found in any dependent variable in the two groups (see Table 7S in SM 4).

---

[4] The correction procedure employed a Generalized Linear Model approach to regress out the scores of each lying profile factor (i.e., LIE_Ability, LIE_Contextuality, LIE_Frequency, LIE_Negativity) for social desirability effects (i.e., SDE and IM). The adjusted scores were calculated using the *adjust* function from the *datawizard* package in Rstudio.

**FIGURE 3**
*Radar plot of the average values at the four* lying profile *factors in the trickster and virtuous cluster.*

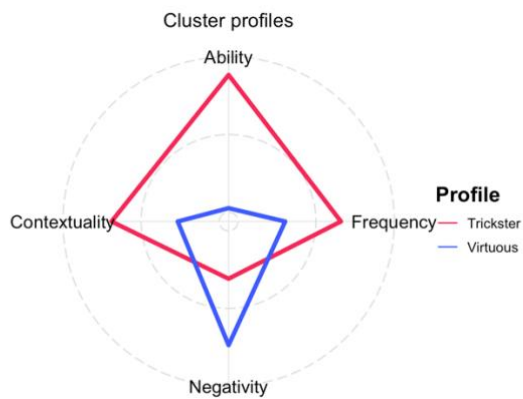

*Note*. The scores at the lying profile factors are corrected for Social Desirability.

## Discussion

The present study aimed to explore a more nuanced yet more ecological form of deception, also known as embedded lies. We provide a new dataset of embedded lies collected in a within-subjects design, providing data at the statement and the individual level to foster psychological research on linguistic and individual differences associated with embedded lies. By collecting a dataset that was sufficiently large to perform predictive modelling, we resorted to supervised machine learning and language models to classify statements as completely truthful or with embedded lies. Furthermore, our study explored the linguistic properties of embedded lies by leveraging automated NLP techniques.

---

[5] The Hopkins' H statistic checks whether the data is appropriate for clustering. We can reject the null hypothesis and conclude that the dataset is significantly clusterable when H < 0.5. A value for H lower than 0.25 indicates a clustering tendency at the 90% confidence level [64]. The H statistic was computed using the *check_clusterstructure* function from the *performance* package in Rstudio.

**The nature of embedded lies**
Our findings suggest that participants used, on average, five embedded lies in their statements to achieve a predefined deception goal. About 1/3 of the length of deceptive statements were embedded lies. Similar figures are reported elsewhere for embedded lies in faked opinions about friends (37%) [2]. As for the source of embedded lies, most participants relied, whether directly or indirectly, on their personal experiences (46.27%), while a smaller percentage used their imagination (33.86%) or drew from others' experiences (14.95%). This finding supports the notion that liars often integrate elements of truth into their lies to enhance plausibility, making the detection of deception more difficult [2]. A realistic deceptive statement (i.e., one with embedded lies rather than a full-blown deceptive narrative) can thus be typified as one that consists of about 2/3 of truthful information and 1/3 of embedded lies, which are most likely to be derived from personal experience.

**Detecting embedded lies**
To assess whether statements containing embedded lies can be differentiated from truthful ones, we trained several machine learning models. The results showed that embedded lies present a significant challenge for deception detection due to their incorporation of truthful elements. Specifically, the highest performance of a language model with competitive capabilities (Llama-3.1-8B) [62], that we fine-tuned, reached 64% accuracy, in line with commonly reported performances in previous research [30 ,65-67]. Notably, a language model published elsewhere - with a reported accuracy of 79.31% in detecting fabricated statements across different contexts [22] - dropped to 56% accuracy when applied to our study. An explanation for that drop could be attributed to overfitting, due to related works showing deception classifiers dropping remarkably when tested on new samples [15]. However, we argue this was not the case. In the original study, the detection rate (i.e., the recall) for truthful (81%) and deceptive statements (78%) was balanced. In contrast, in our study, the deception language model showed a recall of 76% for truthful statements, similar to that of the original study, but a remarkable drop to 35% for deceptive statements. This drop indicates that the struggle was mainly in the detection of embedded lies, which were often misclassified as truthful statements (here: 65% of embedded lies were misclassified as truthful, vs. 22% in the original study). If it were a matter of overfitting, we would have also expected a remarkable drop in the recall of truthful statements. However, this decline was not observed, which indicates that while the deception model was able to resort to what was learnt during the training phase to classify new samples of truthful statements correctly, it was unable to do so for the deceptive ones. We argue this was attributed to the fact that the deception involved was different (here: embedded lies vs. fabrication in the original study). Moreover, the explainability analyses on the Llama-3 and deception language model provided further evidence for the notion that the more nuanced the embedded lies are, the harder they are to detect.

Finally, when employing other common approaches, typically employed to detect deception (i.e., ML models trained on BOW representation, LIWC features, and embeddings), performance was significantly better than chance – albeit reaching just 55% to 62% accuracy. Other fine-tuned language models (here: distilBERT and FLAN-T5 base), were no more effective in performing the task. Altogether, these findings indicate that the challenge in identifying embedded lies stems from their resemblance to truthful statements and, as the degree of fabrication increases, the classification process becomes more straightforward.

**Textual properties of embedded lies**

In addition to individual differences among liars, we examined linguistic properties of embedded lies. Linguistic analysis using psycholinguistic variables, and a deception-specific set of variables (DeCLaRatiVE) revealed several differences between truthful statements and those with embedded lies. Truthful statements contained a larger proportion of social references, while deceptive statements tended to include more references to cognitive processes, such as memory-related words. However, these effect sizes were small, and some linguistic features, such as emotional words and text statistics, exhibited ambiguous patterns, sometimes indicative of deception, and sometimes indicative of truth (e.g., words related to anger being more indicative of truthfulness and those related to sadness being indicative of deception). Similarly, the DeCLaRatiVE analysis suggested that the sole difference between truthful and deceptive statements was in cognitive load indicators (i.e., word and syllable counts), but these findings may have been influenced by the experimental instructions, which encouraged participants to add more details to achieve their deceptive goals. The absence of significant effects for other DeCLaRatiVE variables, especially those from the Reality Monitoring and Verifiability Approach, is in line with previous studies showing that truthful statements did not appear to contain more details compared to embedded lies [2, 33].

When we zoomed in on the event level, we found significant differences in LIWC variables only in four out of eleven events and in DeCLaRatiVE variables in nine out of eleven events. Altogether, these findings suggest that while there are some discernible differences between truthful and deceptive statements, these differences are often subtle and context-dependent.

A term frequency analysis of n-grams underscored the difficulty of detecting deception through word usage when embedded lies are involved. In nine out of eleven events, we found negligible effects, with only one or two significant *n*-grams per event (e.g., "pain" and "surgery" as significant n-grams in deceptive statements for the event "Being hospitalized and undergoing surgery") and with small effect size, highlighting the subtle nature of embedded lies. This supports previous findings that verbal detection remains challenging, particularly when lies are carefully embedded within otherwise truthful narratives [2, 68]. In addition, this overlap can be attributed to the within-subject design employed for this study, which eliminated any potential linguistic confounders derived from having different participants writing about the same task under two conditions (honest vs. deceptive), typical of between-subjects studies.

**Individual differences in embedded lies**

We further investigated individual differences in the nature of embedded lies. We found gender playing a role in the way individuals self-rated the deceptiveness of their embedded lies, with females scoring higher in deceptiveness than males - albeit with small effect sizes. Age also played a role with older participants being more openly deceptive in their statements.

In terms of lying attitude, the results of the cluster analysis were slightly different from the original paper [49]. We identified only two, rather than three, clusters of liars that resembled the original *virtuous* and *trickster* cluster. Specifically, the virtuous cluster was mainly characterised by a strong aversion to deception, while the tricksters tended to lie more frequently, to perceive themselves as good liars, and to adapt their lying behaviour to the context. However, despite this clear distinction, no significant differences were reflected in their behaviour and, specifically, in the absolute and standardised number of embedded lies, as well as in their deceptiveness and centrality scores. A possible explanation for why the difference in the lying attitude was not reflected in the lying behaviour (i.e., in the number of embedded lies) might be that all participants were instructed to rewrite the statement deceptively by adding embedded lies, and this might have reduced the variability in their responses.

**Moving forward on embedded lies**

With this paper, we sought to move the dial towards embedded lies by presenting a dataset of 2084 statements (i.e., truthful vs. deceptive with embedded lies) about past autobiographical events to spark renewed research interest in embedded lies in verbal deception detection. We focused on autobiographical memories due to their relevance in forensic contexts, where the credibility of witnesses' and suspects' statements is assessed and often centred on autobiographical events. Specifically, this dataset offers the possibility to investigate whether traditional theoretical frameworks of deception (e.g., the use-the-best heuristic [69], the verifiability approach [6], etc.), as well as theories relying on manual coding (e.g., the role of complications, common knowledge details, or self-handicapping strategies [36]), work well when applied to embedded lies.

Moreover, it offers a sufficiently large number of statements to conduct meaningful computational analysis and provides granular information on the statement level, including annotations of embedded lies together with rating scores for their deceptiveness, centrality and source of information. By leveraging computational analysis for predictive modelling, this dataset can be employed for a sequence classification task, allowing for the prediction of how and where lies are embedded within truthful narratives, or for a regression task to quantify the extent of deception (e.g., the number of embedded lies) in a given statement.

Furthermore, it offers the possibility of fostering research on specific contexts of deception by providing statements for eleven categories of events deemed relevant for lying, such as exaggerated insurance claims. Finally, this dataset should enable researchers to study individual differences in deception and, specifically, the use and forms of embedded lies by providing demographic variables and personality-related measures, such as attitudes towards lying and social desirability, as well as memory-related measures about the event (i.e., how in the past; how frequently and how well it is remembered; how important it is; which emotional tone the event has). Understanding how context and individual differences influence the nature and frequency of embedded lies would foster psychological knowledge of deception and provide insights for more context- and individual-sensitive deception detection techniques.

**Limitations and future outlooks**

Despite this study's aim to overcome known limitations related to deception detection research (e.g., focusing on fabrication, use of between-subjects designs, and small sample sizes), this study comes with its own limitations. First, embedded lies were both self-reported and self-annotated by the participants, which may have led to subjective interpretations of what constitutes an embedded lie. This variability among participants could reduce the consistency and reliability of the data. However, in our analysis, the number of embedded lies was standardized by computing the ratio of words in embedded lies to the total number of words in the whole deceptive statement. This procedure ensures that the results are not influenced by individual interpretations of what a unit of embedded lie is. Therefore, we recommend future researchers adopt this or other forms of standardization (even during the data collection process) to ensure consistency in data. Second, while the dataset covers eleven distinct events, focusing the investigation on individual events may result in smaller sample sizes, limiting the statistical power and the ability to conduct predictive analyses within specific events. Finally - and in contrast to the study design employed by Markowitz [2] - we conceptualised truthful statements as entirely truthful, while deceptive statements were situated on a continuum ranging from embedded lies to completely fabricated statements. While it is reasonable that individuals may occasionally offer partial truths, it is also feasible to convey completely truthful statements. Consequently, we opted to narrow our focus of investigation by contrasting completely truthful statements with varying degrees of embedded lies. A potential avenue for future research could involve incorporating partial truths, as Markowitz did in his design [2], or alternatively, having three versions of the statement: truthful, embedded lies, and fully deceptive.

**Conclusion**

In this paper, we presented a novel dataset as a resource to encourage research on embedded lies in verbal deception detection. The analysis of individual differences and linguistic properties, as well as the results from predictive modelling and explainability analysis, highlighted how the unique challenge in detecting embedded lies stems from their nuanced nature and resemblance to truthful statements.

# Supplementary Material - 1

## Instructions

**TABLE 1S**
*Full instructions provided to participants when the order of presentations of conditions was first truthful and then deceptive.*

| Truthful condition | Deceptive condition |
|---|---|
| *Your task is to write about the event* **"Being involved in a car accident"** *twice. For now, provide a* **completely truthful statement**. *Then, you are going to write an alternative version of the same event following additional instructions. This means that* **for now** *you should provide a* **detailed, truthful account of that event.**<br><br>*Make sure to use correct spelling and grammar and separate your sentences with punctuation.*<br>*Describe what happened, who was involved, where and when it took place, and why it was memorable to you. Your statement should be at least 300 characters.*<br><br>**IMPORTANT:**<br>*We are aware that AI-assistant tools (e.g., ChatGPT) are increasingly used for tasks on Prolific. Please do not use it for this task. We seek to understand how humans write these statements. If you feel unable to do the task, please leave this spot open for others.*<br><br>*Write your statement about the event* **"Being involved in a car accident"** *here:"* | *In the previous task, you wrote about* **"Being involved in a car accident"**<br><br>*This time you have to* **lie about the event** *you just described in order to obtain a benefit or avoid a loss. Specifically, you have to write an alternative version of the story about* **"Being involved in a car accident"** *in which you are deceptive to* **increase the claimed amount of damage you received to get more money.**<br><br>*Specifically,* **the event should essentially be the same, but you have to fabricate details**. *Afterward, we are going to ask you in which* <u>specific part of the story</u> *you lied and how.*<br><br>*Now,* **write the deceptive version of your story about "${e://Field/loopEvent}".**<br>*Your statement should be at least 300 characters.* <u>Please don't mention in any way that you are lying in this statement.</u><br><br>**IMPORTANT:**<br>*Try to be as convincing as possible. A researcher who is an expert in verbal-lie detection will evaluate your statement. If the experimenter would consider your statement as credible, you will have the chance to win an* **extra 50£ compensation** *by participating in a draw.*<br><br>*We are aware that AI-assistant tools (e.g., ChatGPT) are increasingly used for tasks on Prolific. Please do not use it for this task. We seek to understand how humans write these statements. If you feel unable to do the task, please leave this spot open for others.* |

**TABLE 2S**
*Full instructions provided to participants when the order of presentations of conditions was first deceptive and then truthful.*

| Deceptive condition | Truthful condition |
|---|---|
| *You are going to write about the event* **"Causing a car accident"** *twice.*<br><br>*Now you have to* **lie about the event** *in order to obtain a benefit or avoid a loss. Specifically, you have to write an alternative version of the story about* **"Causing a car accident"** *in which you* **describe the event so that it's not your fault.**<br><br>*Specifically,* **the event should essentially be the same but you have to fabricate details**. *Afterward, we are going to ask you in which <u>specific part of the story</u> you lied and how.*<br><br>*Now,* **write the deceptive version of your story about "Causing a car accident" using the fabrication strategy.**<br>*Your statement should be at least 300 characters. <u>Please don't mention in any way that you are lying in this statement.</u>*<br><br>**IMPORTANT:**<br>*Try to be as credible as possible because the experimenter is an expert in verbal-lie detection and will read your statement to evaluate it as credible or not. If the experimenter would consider your statement as credible, you will have the chance to win an* **extra 50£ compensation** *by participating in a draw.*<br><br>*We are aware that AI-assistant tools (e.g., ChatGPT) are increasingly used for tasks on Prolific. Please do not use it for this task. We seek to understand how humans write these statements. If you feel unable to do the task, please leave this spot open for others.* | *In the first task we asked you to lie about your experience with* **"Causing a car accident".**<br>*Now we ask you to provide the* **truthful version of the same story.**<br><br>*This means you should provide a* **complete truthful statement** *of what happened, without <u>omitting relevant information or adding made-up details</u>, as you may have done before!*<br><br>*Make sure to use correct spelling and grammar and separate your sentences with punctuation.*<br>*Your statement should be at least 300 characters.*<br><br>**IMPORTANT:**<br>*We are aware that AI-assistant tools (e.g., ChatGPT) are increasingly used for tasks on Prolific. Please do not use it for this task. We seek to understand how humans write these statements. If you feel unable to do the task, please leave this spot open for others.*<br><br>*How did things really turn out?*<br>*Now, re-write the truthful version of your statement here. Describe what happened, who was involved, where and when it took place, and why it was memorable to you.* |

# Supplementary Material - 2

## Descriptive statistics

Here we report the descriptive statistics (*M, SD,* Median) of the memory-related variables associated with the events, such as the time elapsed since the event occurred (in months), frequency of recollection, importance, accuracy of the recollection, and emotional valence (Table 3S).

**TABLE 3S**
*Descriptive statistics (M, SD, Median) of participants' responses on the 5-point Likert scale regarding the time elapsed since the event occurred (in months), frequency of recollection, importance, accuracy of the recollection, and emotional valence of the event.*

|  | M | SD | Median |
|---|---|---|---|
| **Time** | | | |
| "*how long ago did the event happen?*" | 8.80 | 6.69 | 7 |
| **Recollection** | | | |
| "*how often do you think or talk about this event?*" | 2.58 | 1.09 | 2 |
| **Importance** | | | |
| "*how important is this event to you?*" | 3.28 | 1.34 | 4 |
| **Accuracy** | | | |
| "*how well do you remember this event?*" | 4.04 | 0.92 | 4 |
| **Valence** | | | |
| "*how would rate this event in emotional terms?*" | -0.19 | 0.7 | -0.5 |

# Correlational analysis

In Figure 1S we show the significant Spearman's rank correlations between the lying profile and BIDR scales and the dependent variables associated with embedded lies.

**FIGURE 2S**
*Correlation matrix.*

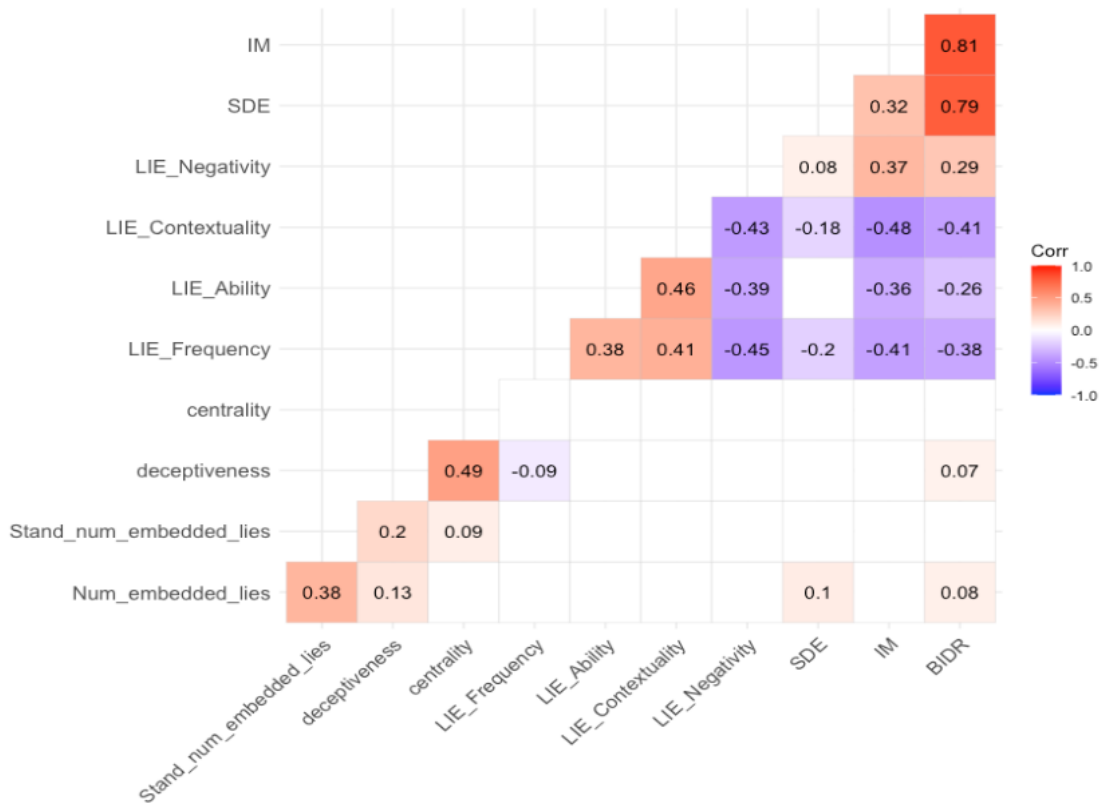

*Note.* Only significant correlations with p < .05 that survived FDR correction for multiple comparisons are reported.
[a]Num_embedded_lies = number of embedded lies
[b]Stan_num_embedded_lie = standardized number of embedded lies
[c]BIDR = Balanced Inventory of Social Desirability Responding scale
[d]SDE = self-deception enhancement subscale of BIDR
[e]IM = impression management subscale of BIDR

## Individual differences

After excluding participants who preferred not to express their gender, expired, or revoked their consent to show gender-related data ($n = 6$), we tested for gender differences in the variables of interest. Specifically, results from a permutation t-test ($n_{perm} = 9.999$) revealed a significant difference between males ($M = 4.68$, $SD = 2.29$) and females ($M = 5.26$, $SD = 3.39$) in the absolute number of embedded lies ($diff = 0.58 \pm 0.20$, $p = 0.003$, $d = 0.18$ [0.06, 0.31]) but not in the standardized number of embedded lies ($diff = 0.01 \pm 0.01$, $p = 0.41$, $d = 0.05$ [-0.07, 0.17]). Therefore, these findings suggest that the difference in gender was mainly driven by statements' length.

There was another gender difference for the average deceptiveness scores ($diff = 0.11 \pm 0.05$, $p = 0.03$, $d = 0.14$ [0.02, 0.26]), with females ($M = 3.98$, $SD = 0.79$) reporting higher values than males ($M = 3.88$, $SD = 0.77$), but not for the average centrality scores ($diff = 0.01 \pm 0.05$, $p = 0.85$, $d = 0.01$ [-0.11, 0.14]). Using Spearman's rank correlations, we found only a small but significant positive correlation between age and deceptiveness ($rho = 0.075$, $S = 172823388$, $p = 0.015$). However, we found no significant correlation between age and the absolute number of embedded lies, the standardized number of embedded lies, as well as age and centrality scores (Table 4S).

**TABLE 4S**
*Spearman's rank correlations between age and the dependent variables.*

| Variables | rho | S | p | Significance ($p < .05$) |
|---|---|---|---|---|
| Age - Absolute no. of embedded lies | -.046 | 178303408 | .137 | No |
| Age - Standardized no. of embedded lies | .002 | 186940849 | .99 | No |
| Age - Deceptiveness | .075 | 172823388 | .015 | Yes |
| Age - Centrality | .03 | 180713689 | .28 | No |

# Supplementary Material - 3

## Machine-learning classification

State-of-the-art ML models were employed in a classification task to distinguish truthful from deceptive statements with embedded lies. We adopted two approaches: a simpler approach where a Random Forest (RF) model was trained on extracted linguistic features, and a more sophisticated approach that involved fine-tuning pre-trained language models.

### Random forest models

The simpler approach consisted of training four Random Forest (RF) models in a binary classification task using as features a Bag of Words (BOW) [59] representations, LIWC variables [11], DeCLaRatiVE variables [22], and GPT-embeddings[6]. RF is an ensemble learning technique that leverages multiple decision trees in the training phase to then select as a final output the most frequent prediction from the individual trees. Below, we provide the details on how we proceeded for the feature extraction:

- **BOW:** we used a bag-of-words representation of unigrams, bigrams, and trigrams to train the model. The bow model was applied on preprocessed text. Preprocessing included lowercasing, lemmatization and the removal of stop-words. We then included only *n*-grams that were present at least 5% of times across documents to exclude rare-words. This bow representation consisted of a vector of length 158.
- **LIWC:** we used the LIWC-22 software to extract 117 syntactic and semantic features from raw text. All features were included in the training phase.
- **DeCLaRatiVE:** we followed the procedure described in [22] and we extracted 26 linguistic features associated with four theoretical frameworks of deception (i.e., Distancing, Cognitive Load, Reality Monitoring, and Verifiability approach). All features were employed to train the model.
- **GPT-embeddings:** the embedding representation was extracting using the OpenAI embeddings models[7]. Specifically, we employed the *text-embedding-3-large* model and, following OpenAI guidelines, we specified a vectorial dimension of 256 using the *dimension* parameter. This allowed us to have a meaningful statement representation without falling in the dimensionality curse (i.e., when the number of features exceeds the number of observations).

The training-test procedure employed a nested cross-validation (NCV) framework. Specifically, it consisted of an inner loop repeated across 10 folds for hyperparameter optimization and an outer loop across 10 folds for model performance evaluation. The hyperparameter optimization was conducted through Grid Search. Once, the best hyperparameter combination was identified in the inner loop, it was then used to train the model on the entire training set in the outer loop. Models' performance was evaluated in terms of accuracy, precision, recall and F1 score.

---

[6] https://platform.openai.com/docs/guides/embeddings/
[7] https://platform.openai.com/docs/guides/embeddings/embedding-models

**Fine-tuning language models**

The value of leveraging language models lies in two key areas: the robust numerical representation of natural language learned during the pre-training phase and the ability to adapt the model to a downstream task with minimal fine-tuning of the parameters in the final layer(s), without altering the underlying architecture. Fine-tuning can be accomplished through further training on task-specific data, which improves the model's capacity to generate coherent and contextually relevant text that aligns with the target task.

To assess models' performance in a robust manner we conducted a 5-fold cross validation, ensuring that both truthful and deceptive statements from the same participants were either present in training test or in test set. This procedure was employed to avoid information leakage and biased performance metrics. Models' performance was assessed in terms of accuracy, precision, recall and F1 score. For our analysis, we tested the performance of fine-tuned versions of distilBERT [60], FLAN-T5 base [21], and Llama-3-8B [61]. All language models, with the exception of the deception language model, were freely available through Hugginface platform.

DistilBERT is a smaller, faster, and cheaper version of the original BERT base model. It was trained by distillation meaning that it was trained to predict the same probabilities as the original BERT model (https://huggingface.co/docs/transformers/en/model_doc/distilbert ). In the present study, distilBERT was fine-tuned using the following configuration of parameters: learning rate 5e-5; weight decay coefficient: 0.01; batch size: 32; number of epochs: 3.

FLAN-T5 is a text-to-text general model developed by Google researchers and capable of solving many NLP task, such as sentiment analysis, question answering, and machine translation (https://huggingface.co/docs/transformers/model_doc/flan-t5). Among the several versions available we employed the FLAN-T5 base, which was fine-tuned with the following configuration: learning rate 5e-5; weight decay coefficient: 0.01; batch size: 2; number of epochs: 3.

Llama-3 model is the most refined version of Llama models, i.e., an open-source collection of foundation language models, developed by Meta AI (https://huggingface.co/docs/transformers/en/model_doc/llama3 ). This generation of Llama models demonstrated state-of-the-art performance on a wide range of benchmarks and showed improved reasoning. We employed the version with eight billion of parameters (Llama-3-8B), which was fine-tuned with a quantized low rank optimization (QLoRA) procedure and the following configuration: learning rate 1e-4; weight decay coefficient: 0.01; batch size: 2; number of epochs: 3.

The deception language model is a fine-tuned version of a FLAN-T5 base model to classify deceptive statements. In the original study, the deception language model was fine-tuned in three datasets encompassing 2500 personal opinions, 5506 autobiographical memories, and 1640 future intentions, reaching 79.31% accuracy [22]. For this study, the deception model was employed as it is to predict deception in our dataset without further fine-tuning.

# Exploratory explainability analysis

Here, we report the exploratory explainability analysis we conducted on the Llama-3 model and deception language model in terms of correlations between deception class probabilities and embedded lies-dependent variables (Table 5S) and differences between correct and incorrect classifications in those dependent variables (Table 6S).

**TABLE 5S**
*Spearman's rank correlations between class probability for deceptive statements and the dependent variables in the Llama-3 and deception language model.*

| Model | Variables | rho | S | p |
|---|---|---|---|---|
| Llama-3-8B model | Absolute no. of embedded lies | .10 | 170216978 | .0009* |
| | Standardized no. of embedded lies | .10 | 170565831 | .001* |
| | Deceptiveness | .05 | 180010090 | .10 |
| | Centrality | .05 | 180335919 | .11 |
| Deception language model | Absolute no. of embedded lies | .09 | 171758230 | .004* |
| | Standardized no. of embedded lies | .01 | 185868739 | .645 |
| | Deceptiveness | -.02 | 191376556 | .630 |
| | Centrality | -.03 | 193262992 | .421 |

*Note.* Positive correlations mean that the class probability of being deceptive (range 0.5 – 1.0) is higher when the dependent variable of interest is higher.
* $p < .01$

**TABLE 6S**
*Embedded lies and associated measures between correct and incorrect classifications in the Llama-3-8B and deception language model.*

| Model | Variables | M (SD) Correct | M (SD) Incorrect | diff (SD) | p | d | 95% CI |
|---|---|---|---|---|---|---|---|
| Llama-3-8B model | Absolute no. of embedded lies | 5.31 (3.39) | 4.43 (2.83) | 0.88 (0.22) | .0001* | 0.27 | 0.14, 0.40 |
| | Standardized no. of embedded lies | 0.34 (0.21) | 0.29 (0.19) | 0.04 (0.01) | .0005* | 0.22 | 0.09, 0.35 |
| | Deceptiveness | 3.95 (0.80) | 3.92 (0.76) | 0.03 (0.05) | .6238 | 0.03 | 0.10, 0.16 |
| | Centrality | 3.57 (0.81) | 3.50 (0.85) | 0.07 (0.05) | .2212 | 0.08 | -0.05, 0.21 |
| Deception language model | Absolute no. of embedded lies | 4.90 (3.30) | 5.10 (3.22) | -0.20 (0.21) | .3475 | -0.06 | -0.19, 0.07 |
| | Standardized no. of embedded lies | 0.33 (0.21) | 0.32 (0.20) | 0.02 (0.01) | .2192 | 0.08 | -0.05, 0.21 |
| | Deceptiveness | 3.93 (0.83) | 3.94 (0.77) | -0.02 (0.05) | .7446 | -0.02 | -0.15, 0.11 |
| | Centrality | 3.58 (0.87) | 3.53 (0.80) | 0.05 (0.05) | .4156 | 0.06 | -0.07, 0.18 |

*Note.* The table reports means (*M*) and standard deviations (*SD*) for each variable. The mean differences (*diff*), *p*-values, Cohen's *d* effect size, and 95% confidence intervals (CI) for the differences come from a permutation t-test with 9.999 permutations.
* $p < .01$

# Supplementary Material - 4

## Clusters of liars

Following the original procedure in Makowski et al., (2021) [49], we investigated the presence of subpopulations of liars by clustering participants' scores at the four-factor lying profile questionnaire. The only difference with the original procedure was that we used the lying profile scores corrected for social desirability. The correction procedure employed a Generalized Linear Model (GLM) approach to regress out the scores of each lying profile factor (i.e., LIE_Ability, LIE_Contextuality, LIE_Frequency, LIE_Negativity) for social desirability effects (i.e., SDE and IM). The adjusted scores were calculated using the *adjust* function from the *datawizard* package in Rstudio. In our dataset, the agreement method procedure suggested an optimal solution with two clusters and a second solution with three clusters. This final two-cluster solution, that we reported in the main text, reflected the *trickster* and the *virtuous* clusters from the original study.

However, with the aim of replicating the findings of the original study, we also applied the k-means algorithm to compute the three-cluster solution. As shown in Fig. 2S Panel D, the obtained clusters were very different from the original ones (see Fig. 2S Panel A). Specifically, we found a group of participants with very low self-reported lying ability, frequency and contextuality, and strong negative emotions and moral attitudes associated with lying, that closely resembled the *virtuous* cluster (40.88% of the sample). A second group of participants (26.20%), which should reflect the *Average* cluster in the original paper, actually, showed average levels of frequency and contextuality, higher levels of negative attitudes and, unexpectedly, extremely high levels of ability. This different score distribution makes this group less analogous to the original one. The third group, which should reflect the *trickster* (32.92%) in the original paper, was composed by people showing very low levels of negativity, extremely high levels of frequency and contextuality and high levels of ability (but less high than in the original paper).

To investigate whether the correction procedure altered the clustering output, we replicated the same analytical procedure as in Makowski et al., (2021) [49], but this time using the raw lying profile scores. The dataset was deemed suitable for clustering (Hopkins' H = 0.27). However, the method agreement procedure again supported the existence of 2 clusters, as indicated by 12 methods out of 29 (41.38%). For replication issues, we applied the k-means clustering algorithm and compared the results when obtaining two and three clusters, respectively. The former accounted for 37.76% of the total variance of the original data, and the latter accounted for 48.98%.

When grouping participants into two clusters (Fig. 2S, Panel B), we found a group of people resembling the *virtuous* cluster in the original study and a group of people resembling the *Average* cluster, but with a lower level of reported frequency. In contrast, when participants were grouped into three clusters (Fig. 2S, Panel C), we obtained a more similar output, yet we failed to fully replicate the original findings due to the frequency scores being overall lower than in the original study.

**FIGURE 2S**
*Radar plot of the average values of the four lying profile factors in the two- and three-cluster solutions.*

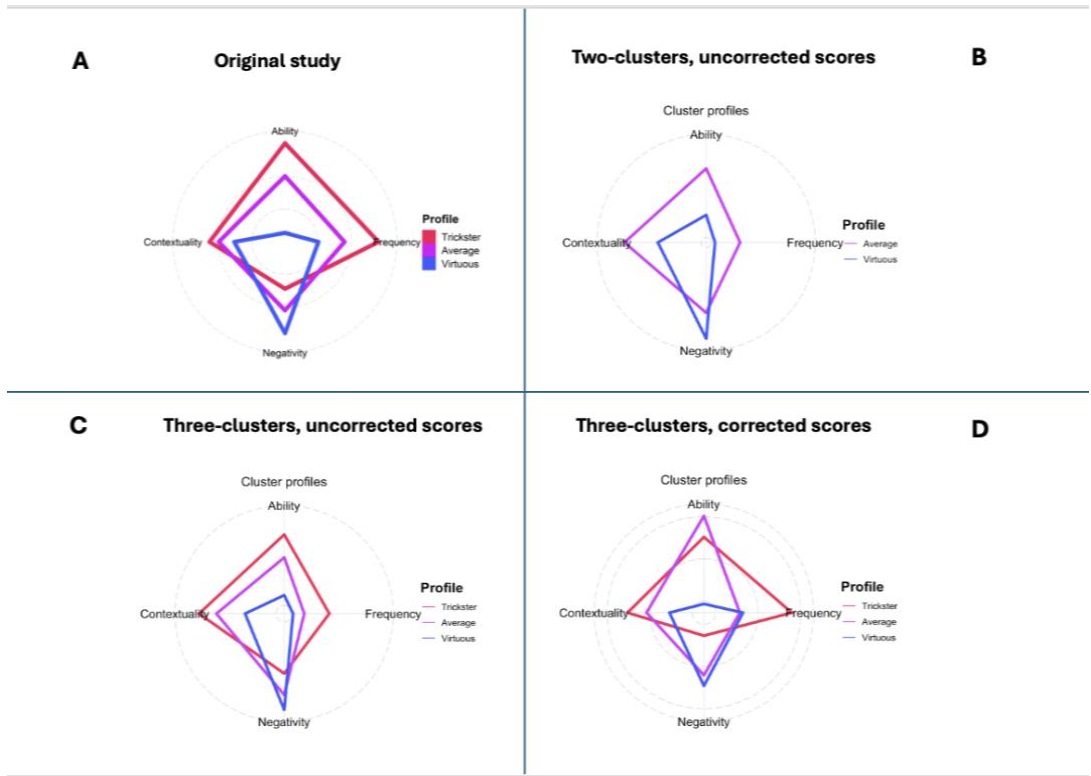

Here we report in Table 7S the results of the paired permutation t-test on the differences between trickster and virtuous in the absolute and standardized number of embedded lies, deceptiveness, and centrality scores. No significant differences were found.

**TABLE 7S**
*Embedded lies and associated measures between trickster and virtuous profiles.*

|  | M (SD) |  | diff (SD) | p | d | 95% CI |
|---|---|---|---|---|---|---|
|  | Trickster | Virtuous |  |  |  |  |
| Absolute no. of embedded lies | 4.91 (3.16) | 5.19 (3.35) | -0.29 (0.20) | .15 | -0.09 | -0.21, 0.03 |
| Standardized no. of embedded lies | 0.32 (0.19) | 0.33 (0.21) | -0.003 (0.01) | .78 | -0.02 | -0.14, 0.10 |
| Deceptiveness | 3.94 (0.81) | 3.94 (0.75) | -0.01 (0.05) | .99 | -0.002 | -0.12, 0.12 |
| Centrality | 3.57 (0.80) | 3.53 (0.84) | 0.04 (0.05) | .46 | 0.05 | -0.08, 0.17 |

*Note.* The table reports means (*M*) and standard deviations (*SD*) for each variable. The mean differences (*diff*), *p*-values, Cohen's *d* effect size, and 95% confidence intervals (CI) for the differences come from a permutation t-test with 9.999 permutations.